\documentclass[letterpaper]{article} 
\usepackage{Anonymous}  
\usepackage{times}  
\usepackage{helvet}  
\usepackage{courier}  
\usepackage[hyphens]{url}  
\usepackage{graphicx} 
\usepackage{natbib}  
\usepackage{caption} 
\usepackage{algorithm}
\usepackage{algorithmic}
\usepackage{newfloat}
\usepackage{listings}

\usepackage{multirow}
\usepackage{amssymb}
\usepackage{amsmath}
\usepackage{graphicx} 

\DeclareCaptionStyle{ruled}{labelfont=normalfont,labelsep=colon,strut=off} 
\floatstyle{ruled}
\newfloat{listing}{tb}{lst}{}
\floatname{listing}{Listing}

\title{Fine-grained Multiple Supervisory Network for Multi-modal Manipulation Detecting and Grounding}

\author {
	Xinquan Yu\textsuperscript{\rm 1},
	Wei Lu\textsuperscript{\rm 1}\thanks{Corresponding authors},
	Xiangyang Luo\textsuperscript{\rm 2}\footnotemark[1]
}
\affiliations {
	\textsuperscript{\rm 1}School of Computer Science and Engineering, Ministry of Education Key Laboratory of Information Technology, Guangdong Province Key Laboratory of Information Security Technology, Sun Yat-sen University, Guangzhou 510006, China\\
	\textsuperscript{\rm 2}State Key Laboratory of Mathematical Engineering and Advanced Computing, Zhengzhou 450002, China\\
	yuxq28@mail2.sysu.edu.cn, luwei3@mail.sysu.edu.cn, 
	luoxy\_ieu@sina.com
}

\begin{document}

\maketitle


\begin{abstract}
	
The task of Detecting and Grounding Multi-Modal Media Manipulation (DGM$^4$) is a branch of misinformation detection. Unlike traditional binary classification, it includes complex subtasks such as forgery content localization and forgery method classification. 
Consider that existing methods are often limited in performance due to neglecting the erroneous interference caused by unreliable unimodal data and failing to establish comprehensive forgery supervision for mining fine-grained tampering traces. 
In this paper, we present a Fine-grained Multiple Supervisory (FMS) network, which incorporates modality reliability supervision, unimodal internal supervision and cross-modal supervision to provide comprehensive guidance for DGM$^4$ detection.
For modality reliability supervision, we propose the Multimodal Decision Supervised Correction (MDSC) module. It leverages unimodal weak supervision to correct the multi-modal decision-making process.
For unimodal internal supervision, we propose the Unimodal Forgery Mining Reinforcement (UFMR) module. It amplifies the disparity between real and fake information within unimodal modality from both feature-level and sample-level perspectives.
For cross-modal supervision, we propose the Multimodal Forgery Alignment Reasoning (MFAR) module. It utilizes soft-attention interactions to achieve cross-modal feature perception from both consistency and inconsistency perspectives, where we also design the interaction constraints to ensure the interaction quality. 
Extensive experiments demonstrate the superior performance of our FMS compared to state-of-the-art methods.

\end{abstract}

\section{Introduction}
In the era of self-media, misinformation has proliferated due to its high popularity and substantial traffic \cite{lazer2018science, wu2019misinformation}. Particularly with the iteration of generative models \cite{goodfellow2020generative, ho2020denoising}, creating misinformation has become easier than ever. For instance, a simple command is sufficient to generate face images or text content with inconsistent attributes \cite{radford2019language, patashnik2021styleclip, kang2023scaling, liu2024deepseek}, which seriously distorts event truths and disrupts public perception.

To address this, researchers have undertaken studies on detecting misinformation. Early methods primarily focused on unimodal data, such as fake text \cite{ma2015detect} and fake images \cite{gupta2013faking}. With the advancement of the internet, multimodal data \cite{shu2020fakenewsnet} has become the mainstream form of information transmission. This has led to the emergence of multimodal misinformation detection. 

\begin{figure}[t]
	\centering
	{\includegraphics[width=\linewidth]{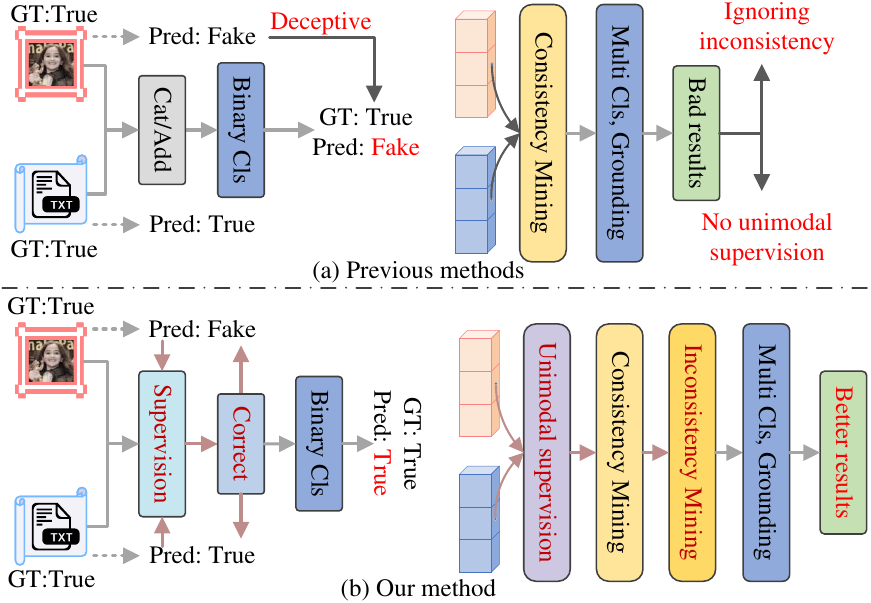}}
	\caption{Comparison of the forgery supervision. (a) Previous methods combine features in a linear manner, which may be misleading due to unreliable modalities. And, previous methods focus on consistency mining. (b) Our method supervises and corrects for unreliable modalities, while enabling comprehensive supervision through unimodal supervision together with cross-modal consistency and inconsistency mining.
	}
	\label{fig0}
\end{figure}

Early research on multimodal misinformation detection \cite{jin2017multimodal, zhou2023multimodal, ying2023bootstrapping} only supports binary classification of authenticity, largely due to the limitations of available datasets. With the proposal of the DGM$^4$ dataset \cite{shao2023detecting}, research has shifted toward finer-grained detection \cite{liu2025idseq}, including forgery content localization and forgery method classification. It mainly includes: HAMMER \cite{shao2023detecting} utilized a two-stage network to handle simple and complex subtasks separately. 
Building on this, HAMMER++ \cite{shao2024detecting} devised a local viewpoint contrast loss to further align cross-modality features. 
Considering the inequality of cross-modal interactions, ViKI \cite{li2024towards} utilized the cross-modality prompt to adaptively aggregate unimodal embeddings, thereby leveraging complementary knowledge in cross-modality embedding.
Since previous approaches primarily mined the manipulated traces from the RGB domain and overlooked the role of the frequency domain, UFAFormer \cite{liu2024unified} utilized the discrete wavelet transform to enrich image representation for the first time. 
To mitigate performance degradation caused by modal competition, MSF \cite{wang2024exploiting} proposed the decoupled fine-grained classifiers to independently guide image and text features. 
In the state-of-the-art method ASAP \cite{zhang2025asap}, the Large Language Model (LLM) was employed to enhance cross-modal alignment.

Although existing methods have yielded promising results, challenges still persist in the fine-grained understanding of forgery traces. As shown in Figure 1(a), this can be attributed to the following reasons, 1) Neglecting the erroneous interference from unreliable modality.
Current methods treat the two unimodal features equally, typically employing concatenation or addition for binary classification. However, not all modality judgments are reliable. When an unreliable modality is present, utilizing such methods may lead to erroneous outcomes due to the disruptive influence of the unreliable unimodal feature. 2) Incomprehensive forgery supervision. Current methods primarily focus on supervising alignment and interaction of cross-modal information from the perspective of consistency, neglecting the role of inconsistent components within cross-modal information in forgery detection. Although the recent approach ASAP \cite{zhang2025asap} utilizes a tamper-guided cross-attention module to constrain interaction, it overlooks the consistency among cross-modal features. Furthermore, few existing methods impose constraints on the mining of forgery traces within individual modalities.

To this end, we propose the Fine-grained Multiple Supervisory Network (FMS), aimed at providing fine-grained constraints for DGM$^4$ to guide the detection of subtle tampering traces. 
Specifically, as shown in Figure 1(b), to mitigate the erroneous interference from unreliable modalities, we design the Multimodal Decision Supervised Correction (MDSC) module. It first imposes weakly supervised constraints on each modality to assess its reliability, and then corrects the unreliable modality through contrastive learning.
Next, to mine tampering traces within individual modalities, we design the Unimodal Forgery Mining Reinforcement (UFMR) module. It first amplifies the gap between real and fake features within a single image or text at the feature-level perspective, and then enhances the gap across multiple images or texts at the sample-level perspective.
Finally, to provide comprehensive forgery supervision, we design the Multimodal Forgery Alignment Reasoning (MFAR) module. It first filters consistent and inconsistent regions guided by the similarity of global features. Subsequently, it utilizes a learnable mask matrix to mine the consistency and inconsistency of cross-modal information through soft-attention interactions, while utilizing interaction constraints to supervise the quality of these interactions.
Our contributions are summarized as follows:
\begin{itemize}
	\item We introduce FMS, a fine-grained multiple supervisory network for multi-modal manipulation detecting and grounding, aiming to provide comprehensive supervision for fine-grained mining of tampering traces.
	\item We devise the MDSC module to mitigate erroneous interference from unreliable modality for the first time.
	\item We propose the UFMR and MFAR modules that provide fine-grained constraints from within and across modalities, respectively. Thereby providing comprehensive forgery supervision for the DGM$^4$ task.
\end{itemize}

\section{Methodology}

\subsection{Overview}
\begin{figure*}
	\centering
	{\includegraphics[width=\linewidth]{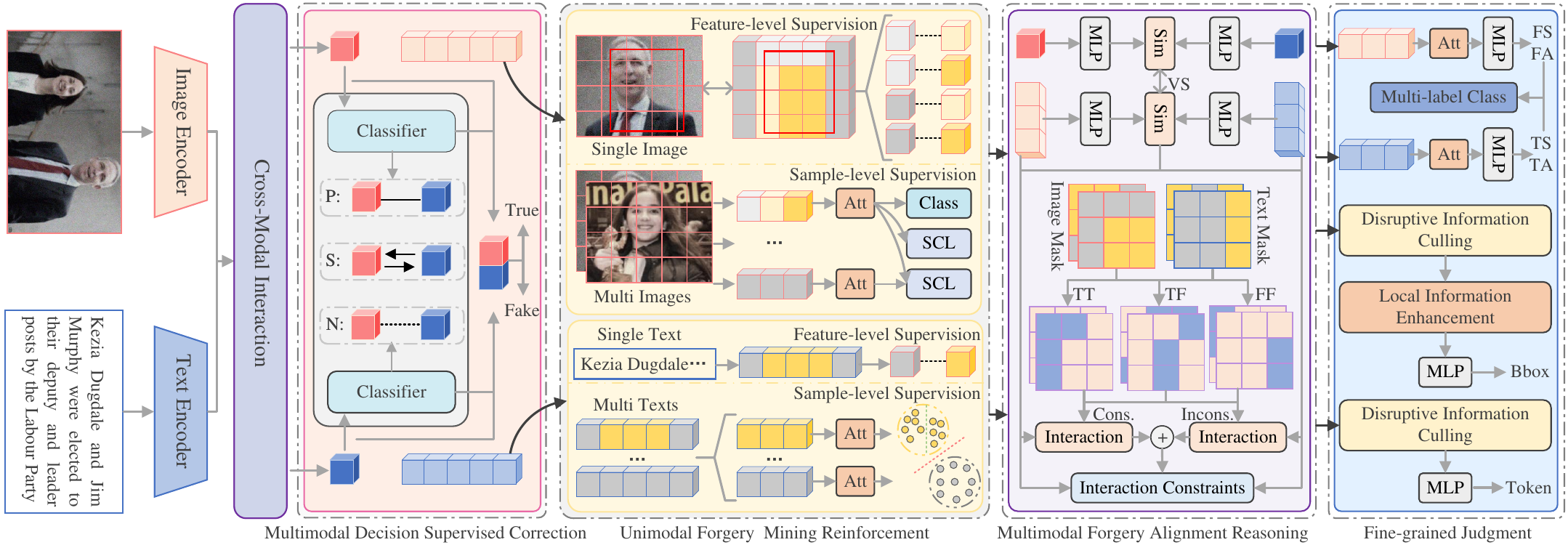}}
	\caption{The overall architecture of proposed FMS. It consists of four components: (1) Multimodal Decision Supervised Correction: this module utilizes unimodal weakly supervised signals to supervise and correct multimodal decisions. (2) Unimodal Forgery Mining Reinforcement: this module amplifies unimodal forgery traces in terms of both feature-level and sample-level losses. (3) Multimodal Forgery Alignment Reasoning: this module ensures the alignment and inference of multimodal forgery traces based on consistency and inconsistency interactions. (4) Fine-grained Judgment: this module utilizes the attention aggregation and disruptive feature culling to achieve fine-grained decision-making.
	}
	\label{fig1}
\end{figure*}

The architecture of our FMS is depicted in Figure \ref{fig1}. Specifically, the multimodal content containing text and images is first fed into the encoder to obtain the initialized embeddings, which are later enhanced by cross-modal interaction. Here, the post-interaction image and text features are denoted as $V = \left [ V_{cls}, V_{pat} \right ] $ and  $T = \left [ T_{cls}, T_{tok} \right ]$, respectively. Then, $V_{cls}$ and $T_{cls}$ are fed into the Multimodal Decision Supervised Correction (MDSC) module, in which the unimodal weak supervision is applied to improve the accuracy of binary classification. Next, $V_{pat}$ and $T_{tok}$ are fed into the Unimodal Forgery Mining Reinforcement (UFMR) module to amplify forgery traces, by calculating the feature-level loss for a single image (or text) and the sample-level loss for multiple images (or texts). Subsequently, $V_{pat}$ and $T_{tok}$ are fed into the Multimodal Forgery Alignment Reasoning (MFAR) module to align multimodal forgery traces, which is achieved by fedding them into consistency and inconsistency interactions, respectively. Finally, the outputs of MFAR are fed into fine-grained judgment module to perform multi-label classification, image and text grounding.

\subsection{Multimodal Decision Supervised Correction}
Binary classification is a subtask in DGM$^4$, aimed at determining whether the given multimodal content is authentic. To achieve this, previous approaches \cite{zhao2024concentrated, jin2024fake, zhang2025asap, li2025unleashing} treat both the image and text modality equally, directly concatenating the two features for multimodal decision-making. However, the accuracy of multimodal decision-making is jointly influenced by both modalities, and once one modality provides misleading information, an unreliable judgment may result. 

Inspired by \cite{zou2023unis}, we leverage the unimodal weak supervision to correct the multimodal decision-making process. 
Specifically, $V_{cls}$ and $T_{cls}$ are first fed into modality-specific classifiers to obtain unimodal predictions. Based on the ground truth, we can compute unimodal prediction loss denoted as $\mathcal{L}^{v}_{uni}$ and $\mathcal{L}^{t}_{uni}$. For example, $\mathcal{L}^{v}_{uni}$ is calculated by 
\begin{equation}
	\mathcal{L}^{v}_{uni} = \mathrm{Mean} \left ( \mathbf{H}\left( \mathrm{MLP} (V_{cls}) \right),y^v_{b-cls}  \right )     
\end{equation}
where $\mathbf{H}$ denotes the cross entropy function, $y^v_{b-cls}$ denotes binary classification labels for image modality.
Note that the unimodal true labels in DGM$^4$ do not necessarily coincide with the binary labels because of the presence of unimodal forgeries. 
Following \cite{zou2023unis}, we divide the features into Positive pairs $\mathbb{P}$, Semi-positive pairs $\mathbb{S}$ and Negative pairs $\mathbb{N}$.
For $\mathbb{P}$, we pull the two closer together, prompting them to become more similar. For $\mathbb{S}$, we pull the ineffective modality closer to the effective modality, prompting the ineffective modality to learn more useful information under weakly supervised signals. For $\mathbb{N}$, we push the two farther apart, making them more dissimilar. 
Thus, we can obtain the multimodal contrastive loss denoted as $\mathcal{L}_{mmc}$,
\begin{equation}
	\mathcal{L}_{mmc} = -\log \frac{\sum_{i\in \{\mathbb{P,S}\}} \exp \left ( \mathrm{sim}\left ( r^v_{i}, r^t_{i} \right ) /\tau   \right )}
	{\sum_{i\in \{\mathbb{P,S,N}\}} \exp \left ( \mathrm{sim}\left ( r^v_{i}, r^t_{i} \right ) /\tau   \right ) }
\end{equation}
where $r^v$ and $r^t$ denote the unimodal representations of $V_{cls}$ and $T_{cls}$ after dimension reduction, respectively. $\mathrm{sim}(\cdot)$ denotes the cosine similarity. $\tau=0.07$ is utilized to control the difference degree.
Finally, we concatenate $V_{cls}$ and $T_{cls}$ to perform binary classification and obtain the classification loss denoted as $\mathcal{L}_{BIC}$. Thus, the overall loss of MDSC can be expressed by
\begin{equation}
	\label{eq3}
	\mathcal{L}_{BIC}^{\star} = \mathcal{L}_{BIC} + \alpha_{1} * \left( \mathcal{L}^{v}_{uni} + \mathcal{L}^{t}_{uni} \right)  + \alpha_{2} * \mathcal{L}_{mmc}
\end{equation}
where $\alpha_1$ and $\alpha_2$ are both trade-off hyperparameters to balance the losses.

\subsection{Unimodal Forgery Mining Reinforcement}
Multimodal detection is essentially an organic combination of multiple unimodal detection tasks, and its performance is closely linked to that of unimodal detection. To this end, UFMR is proposed to amplify the authenticity gap within unimodal features and mine the hidden forgery traces, thereby supporting accurate multimodal detection. It mainly consists two parts: feature-level supervision and sample-level supervision.

\subsubsection{Feature-level Supervision.}
Feature-level supervision takes a single image (or text) as input, aiming to amplify the disparity between patch-to-patch (or token-to-token), thereby enabling better segmentation of real and fake contents. In the case of images, we first divide the ViT-processed image patches into the following categories based on their overlap degree with the forged region,
\begin{equation}
	\gamma_i =\begin{cases} 
		0 & \text{ if } \mathcal{I}_i \in \left [ 0, 0.25\mathcal{P} \right ] \\
		1 & \text{ if } \mathcal{I}_i \in \left [ 0.25\mathcal{P}, 0.5\mathcal{P} \right ] \\
		2 & \text{ if } \mathcal{I}_i \in \left [ 0.5\mathcal{P}, 0.75\mathcal{P} \right ] \\
		3 & \text{ if } \mathcal{I}_i \in \left [ 0.75\mathcal{P}, \mathcal{P} \right ] \\
	\end{cases}
\end{equation}
where $\gamma$ denotes the classified results of each patch. $\mathcal{I}$ and $\mathcal{P}$ denote the intersection area and patch area, respectively. Second, we arrange the above four categories to form four groups $\{\gamma \in \{0,2\}\}$, $\{\gamma \in \{0,3\}\}$, $\{\gamma \in \{1,2\}\}$, $\{\gamma \in \{1,3\}\}$. Then, we extract patch features from each group to amplify their differences,
\begin{equation}
	\mathcal{L}^{v}_{f} = \mathrm{Mean} \left ( W^{v}_{f} \cdot \mathbf{H} \left( \sigma \left( \mathrm{MLP} (V_{pat})_\gamma \right) \right),(y_{pat})_{\gamma}  \right ) 
\end{equation}
where $\mathcal{L}^{v}_{f}$ denotes the feature-level supervision for image modality, $W^{v}_{f}$ denotes the corresponding adaptive weights, which are inversely proportional to the number of corresponding categories in the current group, thereby encouraging the method to assign higher weights to less numerous categories. $\sigma$ denotes the Sigmoid function, $(y_{pat})_{\gamma}$ denotes fine-grained labels of image modality at position $\gamma$. For textual modality, we can similarly compute $\mathcal{L}^{t}_{f}$ in a simpler manner. Finally, we can obtain the overall feature-level supervision loss,
\begin{equation}
	\mathcal{L}_{FS} = \mathcal{L}^{v}_{f} + \mathcal{L}^{t}_{f}
\end{equation}

\subsubsection{Sample-level Supervision.}
Sample-level Supervision takes multiple images (or texts) as input, aiming to amplify the disparity between image-to-image (or text-to-text). In the case of images, we first filter out the real and fake images based on the real labels $y^v$. Next, we execute a standard multi-head attention module \cite{vaswani2017attention} to fuse the features in different patch. Specifically, for real images, we have
\begin{equation}
	\mathcal{G}^v_{r} = \mathrm{MultiAtt} \left ( g^v, V_{pat}^{r}, V_{pat}^{r} \right ) 
\end{equation}
where $g^v$ denotes the global image queries, which is a randomly initialized learnable embedding. $V_{pat}^{r}$ denotes the local features in real images.
For fake images, the situation is slightly different, as we only consider forged regions,
\begin{equation}
	\mathcal{G}^v_{f} = \mathrm{MultiAtt} \left ( g^v, V_{pat}^{f}, V_{pat}^{f}, \mathcal{M}^f_{pat} \right ) 
\end{equation} 
where $V_{pat}^{f}$ denotes the local features in fake images, $\mathcal{M}^f_{pat}$ denotes the attention mask to ignore the real patches in the fake image.
Then, $\mathcal{G}^v_{r}$ and $\mathcal{G}^v_{f}$ are fed into the Supervised Contrastive Learning (SCL) to enlarge the difference between real and fake images \cite{khosla2020supervised},
\begin{equation}
	\begin{split}
		\mathcal{L}^{v}_{s} = &- \mathrm{Mean}\left ( 
		\frac{1}{\left | \mathcal{P}_i \right |} 
		\sum_{j \in \mathcal{P}_i} \log \left( \frac{\exp(\wp_{ij})}{\sum_{k \neq i} \exp(\wp_{ik})} \right)
		\right ) \\
		&+ \lambda \cdot \mathrm{Mean}\left ( 
		\frac{1}{\left | \mathcal{N}_i \right |} 
		\sum_{j \in \mathcal{N}_i} \wp_{ij}	\right )
	\end{split}
\end{equation}
where $\mathcal{P}_i$ and $\mathcal{N}_i$ denote positive and negative sample sets, respectively. $\wp = \mathrm{sim} (\mathcal{G}^v_{r}, \mathcal{G}^v_{f}) /\tau$, $\lambda =0.1$. 
For textual modality, we can also compute $\mathcal{L}^{t}_{s}$ in the same way. 
Thus, the overall sample-level supervision loss is computed by 
\begin{equation}
	\mathcal{L}_{SS} = \mathcal{L}^{v}_{s} + \mathcal{L}^{t}_{s}
\end{equation}

Since fake images are generated through two ways (face swap and face attributes manipulation) in DGM$^4$ \cite{shao2023detecting}, the two leave different forged traces. For this reason, we further refine and supervise $\mathcal{G}^v_{f}$. We first perform
\begin{equation}
	\mathcal{L}^{v}_{m^1} = \mathbf{H} \left( \sigma \left( \mathrm{MLP} (\mathcal{G}^v_{f}) \right) ,y^v_{m} \right)  
\end{equation}
where $y^v_{m}=1$ and $y^v_{m}=0$ denote face swap and face attributes manipulation, respectively. 
Meanwhile, we still utilize the SCL to amplify the difference within $\mathcal{G}^v_{f}$, and obtain $\mathcal{L}^{v}_{m^2}$. Thus, we can compute $\mathcal{L}^{v}_{m} = \mathcal{L}^{v}_{m^1} + \mathcal{L}^{v}_{m^2}$ to assist the subsequent fine-grained face grounding task. For textual modality, the process is the same.

\subsection{Multimodal Forgery Alignment Reasoning}
Apart from enhancing unimodal forgery detection, as mentioned in the previous section, multimodal interaction and alignment are always regarded as a boon for multimodal forgery detection \cite{liu2024mcl,yu2025racmc}. In this view, MFAR is proposed to align the detailed information of the two modalities, thereby supporting forgery detection. 
It mainly consists three layers: global supervision guidance, mask generation learning and soft interaction aggregation.

\subsubsection{Global Supervision Guidance. }
This layer aims to construct consistency and inconsistency matrix while filtering out interfering information, through the supervised guidance of global features.
Specifically, we first calculate the global similarity by
\begin{equation}
	S_g = \mathrm{sim} \left ( \mathrm{MLP} \left ( V_{cls} \right ),  \mathrm{MLP} \left ( T_{cls} \right ) \right ) 
\end{equation}
Second, we calculate the fine-grained similarity denoted as $S_{p-t}$ between $V_{pat}$ and $T_{tok}$ in the same manner.
Then, $S_g$ and $S_{p-t}$ are scaled between 0.5 and 1.5 to provide support for subsequent soft interaction.
Third, we construct consistency matrix denoted as $S_{max}$ utilizing the following constraints,
\begin{equation}
	S_{max} =\begin{cases}
		S_{p-t} & \text{ if } S_{p-t} > S_g \\
		0  & \text{ otherwise }
	\end{cases} 
\end{equation}
while recording the number $n_1$ of selected features. If $n_1 < k$, we will supplement $S_{max}$ by selecting $k-n_1$ additional distinct maximum values. Finally, we can also get cross-modality inconsistency matrix $S_{min}$ in the same way. 

\subsubsection{Mask Generation Learning.}
This layer aims to construct learnable interaction masks to indicate the exact location of cross-modal consistent or inconsistent interactions.
This can be divided into three cases: (1) For the regions where both the image and text are authentic, we apply consistency interaction. (2) For the regions with one authentic and one fake modality, we apply inconsistency interaction. 
(3) For the regions where both are fake, the two interactions mentioned above are applied. 

Taking the first case as an example, we generate an initialized embeddings denoted as ${g}_{tt}$, to learn the masks when both modalities are authentic,
\begin{equation}
	\mathcal{L}_{tt} = \mathrm{Mean} \left ( \mathbf{H} \left( \sigma \left( {g}_{tt} \right) \right), y_{tt} \right ) 
\end{equation}
where $y_{tt} = 1$ only when both modalities are authentic.
Then, we can get ${g}_{ff}$ and $\mathcal{L}_{ff}$ in the same way.
Thus, the consistent masks can be constructed  based on the previously mentioned $S_{max}$,
\begin{equation}
	\chi _{c} = S_{max} \cdot \left ( {g}_{tt} + \alpha_3 \cdot {g}_{ff} \right ) 
\end{equation}
where $\alpha_3=0.5$ is utilized to control the weight of regions that both modalities are fake.

In the same way, the inconsistent masks is computed by
\begin{equation}
	\chi _{ic} = S_{min} \cdot \left ( {g}_{tf} + \alpha_3 \cdot {g}_{ff} \right ) 
\end{equation}
where $g_{tf} = 1 - {g}_{tt} - {g}_{ff}$.
\subsubsection{Soft Interaction Reasoning.}

This module aims to perform fine-grained cross-modal interactions from both consistent and inconsistent perspectives.
In the case of consistency interactions for image modality, we have
\begin{equation}
	\bar{V}_{pat} = V_{pat} + \mathrm{softmax} \left ( \frac{ V_{pat}, (T_{tok})^{T} }{\sqrt{d}} \right ) \cdot \chi _{c} \cdot T_{tok}
\end{equation}
Note that, unlike the hard-mask strategy in traditional attention mechanisms, our strategy not only filters out irrelevant information, but also adaptively controls the attention weights of the selected features, through $\chi _{c}$. Second, we can 
obtain $\widehat{V}_{pat}$ by utilizing $\chi _{ic}$ in the same way. Thus, the post-interaction image features denoted as $\widetilde{V}_{pat}$ can be obtained by 
\begin{equation}
	\widetilde{V}_{pat} = \bar{V}_{pat} + \widehat{V}_{pat}
\end{equation}
Third, we further design interaction constraints, to ensure that the binary classification after interaction is accurate, and the detection performance is improved, compared to without interaction. This process is modeled as
\begin{equation}
	\mathcal{L}^{v}_{ic} = \mathcal{L}^{v}_{ai} + \mathcal{L}^{v}_{ni} + \eta \cdot \Phi( \mathcal{L}^{v}_{ai} -  \mathcal{L}^{v}_{ni})
\end{equation}
where $\mathcal{L}^{v}_{ic}$ denotes the interaction constraints loss. $\eta$ is a hyperparameter to control the punishment intensity. $\Phi(\cdot)$ denotes the ReLU activation function. $\mathcal{L}^{v}_{ai}$ and $\mathcal{L}^{v}_{ni}$ represent the detection loss after interaction and without interaction, respectively. Both of them have a similar calculation process. For example, $\mathcal{L}^{v}_{ai}$ is computed by 
\begin{equation}
	\mathcal{L}^{v}_{ai} = \mathrm{Mean} \left ( W^{v}_{ai} \cdot \mathbf{H} \left( \sigma \left( \mathrm{MLP} \left ( \bar{V}_{pat} \right )  \right) \right), y_{pat} \right ) 
\end{equation}

For textual modality, we can compute $\widetilde{T}_{tok}$ and $\mathcal{L}^{t}_{ic}$ in the same way. Thus, the overall cross-modal interaction loss is 
\begin{equation}
	\mathcal{L}_{CI} = \mathcal{L}_{tt} + \mathcal{L}_{ff} + \mathcal{L}^{v}_{ic} + \mathcal{L}^{t}_{ic}
\end{equation}

\subsection{Fine-grained Judgment}

This module aims to utilize the features extracted above to achieve fine-grained decision-making, including multi-classification, image grounding and text grounding. 

\subsubsection{Multi Classification Task.} Considering that the multi-classification task is an extension of the binary classification task, we utilize the features from binary classification to guide the multi-classification task. Taking image multi classification as an example, we have
\begin{equation}
	V_c = V_{cls} + \mathrm{MultiAtt} \left (V_{cls}, \widetilde{V}_{pat}, \widetilde{V}_{pat} \right) 
\end{equation}
Then, $V^c$ is used to predict face fine-grained manipulation type including FS and FA, thus we can compute image fine-grained classification loss $\mathcal{L}_{m-cls}^v$ following \cite{wang2024exploiting}. For textual modality, we can also get $T^c$ and $\mathcal{L}_{m-cls}^t$ in the same manner. Thus, the overall multi-classification loss is
\begin{equation}
	\label{eq25}
	\mathcal{L}_{MLC}^{\star}=\mathcal{L}_{mlc}^v + \mathcal{L}_{mlc}^t +  \mathcal{L}_m^v + \mathcal{L}_m^t 
\end{equation}
where $\mathcal{L}_m^v$ and $\mathcal{L}_m^t$ are defined in the above section to achieve better supervision.

\subsubsection{Grounding Tasks.} For grounding tasks, we first propose a disruptive information culling layer to reduce the interference from cross-modal information. In the case of image grounding, we execute
\begin{equation}
	\label{eq26}
	V_g = \begin{cases}
		V_{pat} & \text{ if } y^v_{b-cls}==1\,\&\,y^t_{b-cls}==0 \\
		\widetilde{V}_{pat}& \text{ otherwise}
	\end{cases}
\end{equation}
This means that we utilize ${V}_{pat}$ instead of $\widetilde{V}_{pat}$ for image localization only when the image is fake and the text is true, to avoid interference, and vice versa. Second, we utilize a randomly initialized learnable embedding to learn global information with the supervision of $V_{cls}$,
\begin{equation}
	\label{eq27}
	\mathcal{G}^v_{g} = \mathrm{MultiAtt} \left ( g^v_{g}, \mathrm{Cat} (V_{cls},V_g), \mathrm{Cat} (V_{cls},V_g) \right ) 
\end{equation}
Then, $\mathcal{G}^v_{g}$ is feed into the specific classifiers to achieve image grounding. Finally, for textual modality, we can obtain $T_{g}$ and $\mathcal{G}^t_{g}$ in the same way by using Eq. \eqref{eq26} and Eq. \eqref{eq27}, respectively, so as to predict fake text content.

\subsubsection{Full Objectives.} Here we elaborate on all supervisions included in the proposed method. Specifically, there are two aspects of supervision involved, one is designed to improve detection performance, and the other is the original supervision from DGM$^4$ \cite{shao2023detecting}. For the former, we combine all the losses involved,
\begin{equation}
	\mathcal{L}_{our} = \mathcal{L}_{FS} + \mathcal{L}_{SS} + \mathcal{L}_{CI}
\end{equation}
For the latter, we utilize the same supervision function as outlined in \cite{wang2024exploiting}. Note that we enrich two original supervisions in Eq. \eqref{eq3} and Eq. \eqref{eq25} for binary-classification ($\mathcal{L}_{BLC}^{\star}$) and multi-classification ($\mathcal{L}_{MLC}^{\star}$) tasks, respectively. 
By optimizing both, we achieve multi-modal manipulation detection and grounding.

\section{Experiments}
\subsection{Datasets and Metrics}

All experiments in this paper are conducted on the DGM$^4$ dataset \cite{shao2023detecting}, which contains four types of manipulation including Face Swap (FS), Face Attribute (FA), Text Swap (TS), and Text Attribute (TA). In more detail, it has 230,000 image-text pairs comprising 77,426 genuine pairs and 152,574 manipulated pairs. Among the manipulated pairs, there has 123,133 fake images ( 54\% FS and 46\% FA), 62,134 fake texts (70\% TS and 30\% TA) and 32,693 mixed-manipulation pairs that comprises 21\% of all manipulated pairs. 

For model evaluation, we adopt the same metrics as the previous approaches \cite{shao2024detecting, zhang2025asap}. 
Specifically, ACC, AUC and EER are applied to evaluate the binary classification task. 
MAP, CF1 and OF1 are applied to evaluate the fine-grained multi-classification task.
IoUmean, IoU50 and IoU75 are applied to evaluate the image grounding task.
Precision, Recall, and F1-score are applied to evaluate the text grounding task.

\subsection{Implementation Details}

All experiments in this paper are conducted on 8 A100 GPUs. Specifically, for fairness, we first resize the images to $256 \times 256$ and pad the text to a length of 50. Subsequently, following previous studies \cite{wang2024exploiting,li2025unleashing}, we use ViT-B/16 \cite{dosovitskiy2020image} and RoBERTa \cite{liu2019roberta} to extract image and text features, respectively, with the pretraining weights loaded from METER \cite{dou2022empirical}, so as to obtain $V$ and $T$. Additionally, all attention blocks mentioned in the paper consist of 4 layers, with a dropout rate of 0. During the training phase, we use AdamW as the optimizer with the weight decay of 0.02 and learning rate of $1\times 10^{-5}$. Finally, we train for 50 epochs with a batch size of 32, and select the weights from the last epoch for testing.

\begin{table*}[t]
	\setlength{\tabcolsep}{1mm}
	\centering
	\begin{tabular}{cllcccccccccccc}
		\hline\hline
		& \multirow{2}{*}{Method} & \multirow{2}{*}{Reference} & \multicolumn{3}{c}{Binary Cls} & \multicolumn{3}{c}{Multi-label Cls} 
		& \multicolumn{3}{c}{Image Grounding} & \multicolumn{3}{c}{Text Grounding} \\
		& &                       & AUC      & EER$\downarrow$      & ACC      & mAP        & CF1        & OF1       &IoUmean     & IoU50     & IoU75     & Precision    & Recall    & F1 \\ \hline
		\multirow{9}{*}{\rotatebox{90}{\textbf{Entire Dataset}}}
		& CLIP                  & ICML21                & 83.22    & 24.61    & 76.40    & 66.00      & 59.52      & 62.31  & 49.51       & 50.03     & 38.79     & 58.12        & 22.11     & 32.03    \\
		& ViLT                & ICML21                & 85.16    & 22.88    & 78.38    & 72.37      & 66.14      & 66.00  & 59.32       & 65.18     & 48.10     & 66.48        & 49.88     & 57.00   \\
		& HAMMER                  & CVPR23                & 93.19    & 14.10    & 86.39    & 86.22      & 79.37      & 80.37  & 76.45       & 83.75     & 76.06     & 75.01        & 68.02     & 71.35    \\
		& HAMMER++               & TPAMI24               & 93.33    & 14.06    & 86.66    & 86.41      & 79.73      & 80.71  & 76.46       & 83.77     & 76.03     & 73.05        & 72.14     & 72.59   \\
		& UFAFormer              & IJCV24                & 93.81    & 13.60    & 86.80    & 87.85      & 80.31      & 81.48  & 78.33       & 85.39     & 79.20     & 73.35        & 70.73     & 72.02   \\
		& MSF                  & ICASSP24              & 95.11    & 11.36    & 88.75    & 91.42      & 83.60      & 84.38  & 80.83       & 88.35     & 80.39     & 76.51        & 70.61     & 73.44   \\
		& IDseq                  & AAAI25                  & 94.55    & 11.40    & 88.94    & 90.01      & 83.00      & 84.90  & 83.33       & 89.39     & 86.10     & 75.96        & 71.23     & 73.52   \\
		& ASAP                   & CVPR25                & 94.38    & 12.73    & 87.71    & 88.53      & 81.72    & 82.89    & 77.35       & 84.75     & 76.54     & 79.38        & \textbf{73.86}     & \textbf{76.52}   \\
		& FMS (Ours)                    & -         & \textbf{96.46}      & \textbf{9.65}         & \textbf{90.54}          & \textbf{93.43}           & \textbf{86.80}           & \textbf{87.68}	& \textbf{84.82}            & \textbf{91.00}          & \textbf{88.03}          & \textbf{79.43}            & 71.66          & 75.34	\\
		\hline\hline 
	\end{tabular}
	\caption{Comparison with state-of-the-art multi-modal methods on the entire DGM$^4$. The best results are bold.}
	\label{tab1}
\end{table*}

\begin{figure}[t]
	\centering
	{\includegraphics[width=\linewidth]{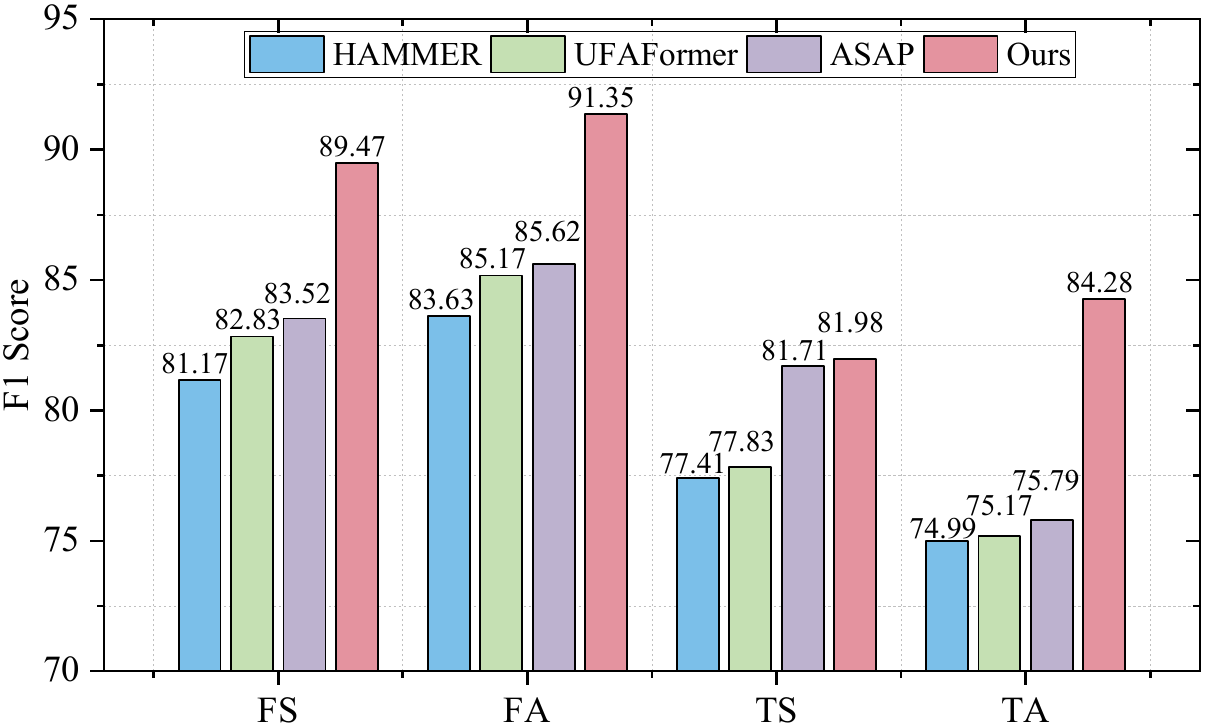}}
	\caption{Comparison about fine-grained classification of forgery methods.}
	\label{fig2}
\end{figure}

\begin{table*}[t]
	
	\centering
	\begin{tabular}{cllcccccc}
		\hline\hline
		& \multirow{2}{*}{Method} & \multirow{2}{*}{Reference} & \multicolumn{3}{c}{Binary Cls} & \multicolumn{3}{c}{Image Grounding} \\
		&	&       & AUC      & EER$\downarrow$      & ACC      & IoUmean     & IoU50     & IoU75     \\  \hline
		\multirow{7}{*}{\rotatebox{90}{\textbf{Image Sub.}}} 
		&TS \cite{luo2021generalizing}                     & CVPR21                & 91.80    & 17.11    & 82.89    & 72.85       & 79.12     & 74.06     \\ 
		&MAT \cite{zhao2021multi}                    & CVPR21                & 91.31    & 17.65    & 82.36    & 72.88       & 78.98     & 74.70     \\ 
		& HAMMER \cite{shao2023detecting}                 & CVPR23                & 94.40    & 13.18    & 86.80    & 75.69       & 82.93     & 75.65     \\
		&HAMMER++ \cite{shao2024detecting}               & TPAMI24               & 94.69    & 13.04    & 86.82    & 75.96       & 83.32     & 75.80     \\
		&ViKI  \cite{li2024towards}                  & IF24                  & 91.85    & 15.92    & 84.90    & 75.93       & 82.16     & 74.57     \\
		&UFAFormer  \cite{liu2024unified}             & IJCV24                & 94.88    & 12.35    & 87.16    & 77.28       & 85.46     & 78.29     \\
		& FMS (Ours)        & -        & \textbf{97.31}        & \textbf{8.21}        & \textbf{91.67}        & \textbf{83.45}           & \textbf{90.60}         & \textbf{87.17}       \\
		\hline\hline 
	\end{tabular}
	\caption{Comparison between our FMS and unimodal methods on the image subset of DGM$^4$. The best results are bold.}
	\label{tab3}
\end{table*}

\begin{table*}[t]
	\centering
	\begin{tabular}{cllcccccc}
		\hline\hline
		& \multirow{2}{*}{Method} & \multirow{2}{*}{Reference} & \multicolumn{3}{c}{Binary Cls} & \multicolumn{3}{c}{Text Grounding} \\
		& &      & AUC      & EER$\downarrow$      & ACC      & Precision    & Recall    & F1      \\ \hline
		\multirow{7}{*}{\rotatebox{90}{\textbf{Text Sub.}}}
		& BERT \cite{devlin2019bert}                   & NAACL19               & 80.82    & 28.02    & 68.98    & 41.39        & 63.85     & 50.23   \\
		& LUKE \cite{yamada2020luke}                   & EMNLP20               & 81.39    & 27.88    & 76.18    & 50.52        & 37.93     & 43.33   \\
		& HAMMER \cite{shao2023detecting}                 & CVPR23                & 93.44    & 13.83    & 87.39    & 70.90        & \textbf{73.30}     & 72.08   \\
		& HAMMER++ \cite{shao2024detecting}               & TPAMI24               & 93.49    & 13.58    & 87.81    & 72.70        & 72.57     & 72.64   \\
		& ViKI \cite{li2024towards}                   & IF24                  & 92.31    & 15.27    & 85.35    & 78.46        & 65.09     & 71.15   \\
		& UFAFormer \cite{liu2024unified}               & IJCV24                & 94.11    & 12.61    & 84.71    & 81.13        & 70.73     & 75.58   \\
		& FMS (Ours)  & -   & \textbf{96.35}        & \textbf{9.67}        & \textbf{89.88}      & \textbf{84.81}            & 71.66         & \textbf{77.68}		\\
		\hline\hline 
	\end{tabular}
	\caption{Comparison between our FMS and unimodal methods on the text subset of DGM$^4$. The best results are bold.}
	\label{tab4}
\end{table*}

\begin{table*}[!t]
	\centering
	\begin{tabular}{ccccccccccccc}
		\hline\hline
		\multirow{2}{*}{Method} & \multicolumn{3}{c}{Binary Cls}                    & \multicolumn{3}{c}{Multi-label Cls}                                & \multicolumn{3}{c}{Image Grounding}              & \multicolumn{3}{c}{Text Grounding}               \\
		& AUC            & EER$\downarrow$ & ACC            & mAP            & \multicolumn{1}{c}{CF1} & \multicolumn{1}{c}{OF1} & IoUmean        & IoU50          & IoU75          & PR.            & Recall            & F1             \\ \hline
		w/o MDSC                & 95.91          & 9.92            & 90.23          & 92.94          & 86.53                   & 87.26                   & 84.00          & 90.26          & 86.96          & 78.00          & 71.36          & 74.53          \\
		w/o UFMR                & 96.07          & 9.79            & 90.27          & 93.22          & 86.56                   & 87.25                   & 81.45          & 89.37          & 80.44          & 78.99 & 71.51          & 75.07          \\
		w/o MFAR                & 95.45          & 9.87            & 90.31          & 93.13          & 86.74                   & 87.50                   & 84.17          & 90.48          & 87.13          & 78.55          & \textbf{71.85}          & 75.05          \\
		Full                    & \textbf{96.46}      & \textbf{9.65}         & \textbf{90.54}          & \textbf{93.43}           & \textbf{86.80}           & \textbf{87.68}	& \textbf{84.82}            & \textbf{91.00}          & \textbf{88.03}          & \textbf{79.43}            & 71.66          & \textbf{75.34} \\
		\hline\hline
	\end{tabular}
	\caption{Ablation study of FMS with different components. PR. represents precision. The best results are bold. }
	\label{tab5}
\end{table*}

\begin{figure*}[!t]
	\centering
	{\includegraphics[width=\linewidth]{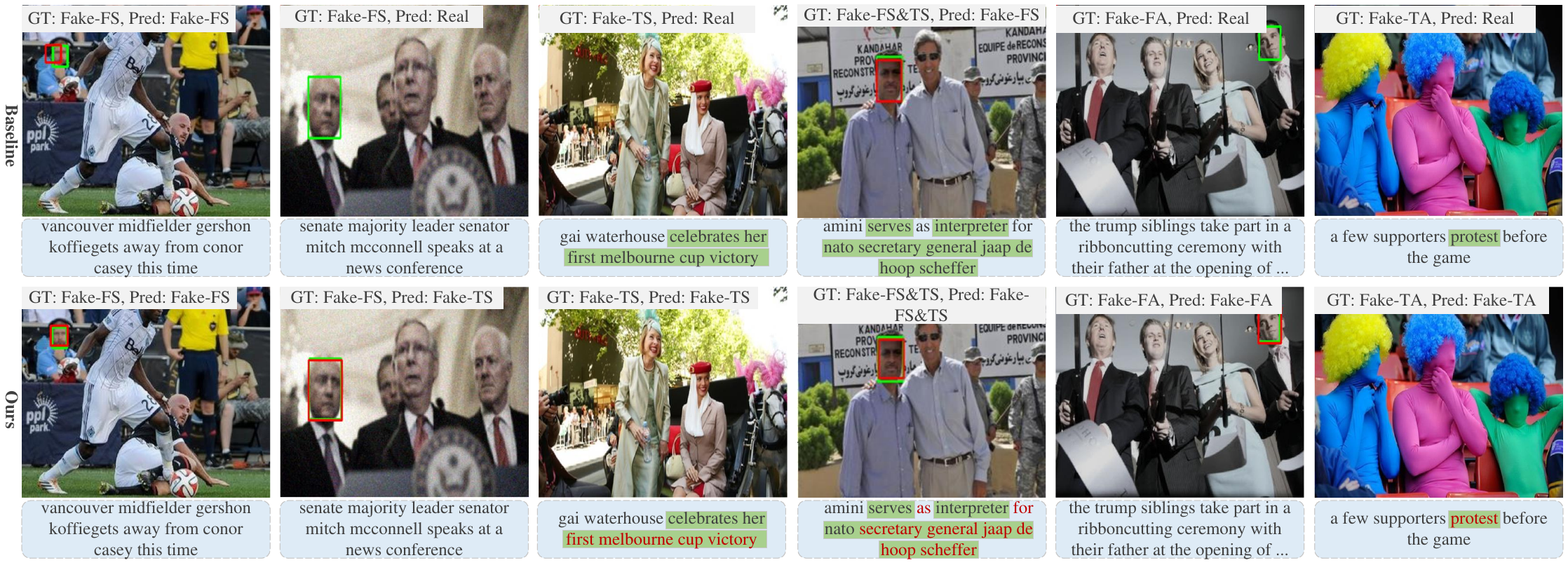}}
	\caption{Visualization of detection and grounding results. Here, red box and text indicate the prediction of manipulated faces and words, while green box and text represent the corresponding ground truth.}
	\label{fig3}
\end{figure*}

\subsection{Comparison with the State-of-the-Art Methods}
\subsubsection{Comparison with Multi-modal Methods.} 
Table \ref{tab1} shows the comparison results between our FMS and eight state-of-the-art methods on the entire DGM$^4$ dataset. 

It can be observed that our FMS outperforms all the comparative methods mentioned above and achieves the highest performance.
Specifically, for the binary classification task, our FMS achieves 96.46\% AUC, improving by 1.35\%--13.24\% over the comparison method, while improving 2.08\% on the latest proposed ASAP \cite{zhang2025asap}.
For the multi classification task, our FMS achieves 93.43\% mAP, improving by 2.01\%--27.43\% over the comparison method, while improving 4.90\% on ASAP.
What’s more, for image grounding task, our FMS attains 84.82\% IoUmean, 91.00\% IoU50 and 88.03\% IoU75. Compared to ASAP, we gains $+$7.47\%, $+$6.25\% and $+$11.49\% on IoUmean, IoU50 and IoU75, respectively. 
For text grounding task, our FMS attains 75.34\% F1, gaining $+$1.82\% over IDseq \cite{liu2025idseq}.
However, our approach is slightly inferior to ASAP because it enriches textual information with LLM.

Besides, we conduct comparative experiments on the fine-grained classification of forgery methods. As shown in Figure \ref{fig2}, our method consistently outperforms HAMMER, UFAFormer and ASAP. For example, for FS, our method gains $+$8.30\%, $+$6.64\% and $+$5.95\% over HAMMER, UFAFormer and ASAP, respectively. For TA, our method gains $+$9.29\%, $+$9.11\% and $+$8.49\% over HAMMER, UFAFormer and ASAP, respectively.
This reflects the effectiveness of fine-grained multiple supervision of FMS.

\subsubsection{Comparison with Unimodal Methods.} 

Similar to previous methods \cite{shao2024detecting, liu2024unified}, we conduct comparisons with unimodal methods on the image and text subsets, respectively. Table \ref{tab3} shows the comparison results for the image subset, where the image is forged. It can be noticed that our FMS attains the best scores with 97.31\% AUC, 91.67\% ACC, 83.45\% IoUmean, 91.00\% IoU50, 87.17\% IoU75, which significantly outperforms the other methods. Compared with MAT \cite{zhao2021multi}, our FMS gains $+$6.00\% AUC, $+$9.31\% ACC, $+$10.57\% IoUmean and $+$12.47\% IoU75.
We consider that this is because MAT only mined tamper traces from the image viewpoint and neglected the text contribution for multimodal detection. In contrast, our approach achieves performance improvement by utilizing multimodal learning.

Table \ref{tab4} shows the comparison results for the text subset, where our method is also the best. Especially in the text grounding task, our method gains $+$27.45\% F1 and $+$34.35\% F1 on BERT and LUKE, respectively. Compared with UFAFormer, our FMS also improves 2.24\% AUC, 5.17\% ACC and 2.10\% F1. 

\subsection{Ablation Study}

To verify the necessity of the proposed components, we removed MDSC, UFMR, and MFAR, respectively. The results are shown in Table \ref{tab5}, where the performance of the full method is the best. Specifically, when removing MDSC, the classification performance decreases significantly, where mAP decreases by 0.49\%. When removing UFMR or MFAR, the grounding performance is also impaired. This sufficiently demonstrates that each of the mentioned components is meaningful for DMG$^4$.

In addition, we devise a baseline, which directly utilizes the features after cross-modal interactions for detection and localization. The visualized comparison results are shown in Figure \ref{fig3}. It can be seen that our method better mines the fine-grained tampering traces. 

\section{Conclusion}
In this paper, we propose a framework named FMS to provide comprehensive supervision for tampering trace mining. 
We first introduce MDSC to assess modality reliability and correct the erroneous interference of unreliable modality.
Second, we develop UFMR to supervise the disparity between real and fake information within unimodal data, thereby enhancing the forgery detection capability of unimodal features.
Finally, we present MFAR to align cross-modal features from both consistency and inconsistency perspectives, thereby facilitating effective interaction between cross-modal information.
Extensive experiments demonstrate that FMS significantly outperforms comparable methods, especially on the image grounding task, where we gain more than 8\% on average over the latest method.

\bibliography{Anonymous}

\end{document}